\documentclass[letterpaper, 10 pt, conference]{ieeeconf}  
\usepackage{amsmath}
\usepackage{graphicx}
\usepackage{subcaption}

\IEEEoverridecommandlockouts                              

\title{\LARGE \bf
Interacting safely with cyclists using Hamilton-Jacobi reachability and reinforcement learning
}

\author{
Aarati Andrea Noronha$^{1}$ and Jean Oh$^{2}$\\
{\small\textit{This manuscript was completed in 2020 as part of the first author’s graduate thesis at Carnegie Mellon University.}}
}

\begin{document}

\maketitle
\thispagestyle{empty}
\pagestyle{empty}

\begin{abstract}
\it In this paper, we present a framework for enabling autonomous vehicles to interact with cyclists in a manner that balances safety and optimality. The approach integrates Hamilton–Jacobi reachability analysis with deep Q-learning to jointly address safety guarantees and time-efficient navigation. A value function is computed as the solution to a time-dependent Hamilton–Jacobi–Bellman inequality, providing a quantitative measure of safety for each system state. This safety metric is incorporated as a structured reward signal within a reinforcement learning framework. The method further models the cyclist’s latent response to the vehicle, allowing disturbance inputs to reflect human comfort and behavioral adaptation. The proposed framework is evaluated through simulation and comparison with human driving behavior and an existing state-of-the-art method.
\end{abstract}

\section{INTRODUCTION}
Existing literature states that the motion of a cyclist is  random, slow and unpredictable when compared to other agents encountered on roadways.
This makes interactions between cyclists and autonomous vehicles particularly challenging. Recent studies have identified various problem areas pertaining to cyclists when it comes to navigation \cite{apasnore2017bicycle, lamondia2012analysis}. They have also   analysed and documented factors that might influence cyclist behaviour and perception. However, there exists a dearth amongst algorithms that are successful at following cyclists, passing cyclists and interacting with them at intersections. Most navigation algorithms tend to be overtly cautious in their interactions and do not offer any guarantees towards optimality or completeness. They also fail at quantifying the safety aspect of their interactions \cite{werneke2015safety}.

In recent years, a heavy emphasis has been placed on safety when it comes to solving optimal control problems. Work on Hamilton-Jacobi reachability analysis has allowed for  solutions to these problems \cite{mitchell2005time}. They guarantee safety by completely barring ego agents from entering user-defined failure zones. But this  would come at the cost of sacrificing the algorithm's ability to reach the goal state within a reasonable time duration. Reinforcement learning algorithms such as deep Q learning have proven useful at providing suitable trade-offs between optimality and completeness for traditional optimal control problems \cite{dhiman2018critical}.

In 2019, Jamie et al. \cite {nachum2017bridging} introduced a  time discounted Hamilton Jacobi Safety Bellman equation used in conjunction with Q-Learning to render reinforcement learning suitable from the perspective of safety. This paper targetted systems with single agents. We built upon this approach and extended it to a system with two agents. 

In summary, our main contributions are a navigation algorithm: 1) that extends the approach proposed by Fisac et al. to a system with two agents, 2) that quantitatively defines the \textbf{cyclist's comfort level } to the autonomous vehicle's state  3) that allows for an autonomous vehicle to interact with cyclists in an \textbf{optimal} manner  in terms of both safety and time taken,  4) that effectively \textbf{quantifies safety}: the ability to pass a cyclist at a reasonable speed and reasonable distance.

\section{Problem Formulation}

Current autonomous driving algorithms are too cautious when passing cyclists on the road resulting in sub-optimal interactions in terms of time taken. Human drivers in contrast often end up being overtly aggressive in their maneuvers. The problem formulation is the same as that defined by Fisac et al. (2019)~\cite{fisac2019} except we introduce the cyclist as a second agent.

Given a start and a goal, the aim is to generate a trajectory $\epsilon$ for the vehicle  that maximizes an objective function  $V:R^n \times [-T,0] \rightarrow R$ which is reflective of both safety and time taken.
\begin{equation}
     V(x) :=\sup_{u(\cdot)}\inf_{t>=0}g(\epsilon(t:x,u(\cdot),d(\cdot)))
 \end{equation}
where $u$ is the control input of the vehicle and $d$ is the disturbance input of the cyclist.
It can be interpreted as a sequence generation problem where we are given the start and the goal and the aim is to generate an appropriate control input $u$ at every time instant $t \in T$ given a disturbance $d$.

The study has three parameters of interest: passing the cyclist whilst maintaining a reasonable distance, passing the cyclist at a reasonable speed, and reaching the goal position in the minimum possible time.

\section{Related Work}

One of the reasons for the failure to interact safely with cyclists is the lack of naturalistic driving data available. There exists an abundance of crash data that documents accidents involving cyclists \cite{dozza2017crash}. But this data offers no clear insight into the state of the system before the crash, environmental factors or the presence of other agents. Few naturalistic databases but the Safety Pilot Michigan database, the inD dataset \cite{inDdataset}  and the Stanford Drone Database effectively track events involving cyclists. The Safety Pilot Michigan database is deployed for the purpose of this study.

Typically optimal control problems of the type above have been tackled in literature using Hamilton-Jacobi reachability \cite{bansal2017hamilton}. Basically the optimal control problem is formulated as a differential game which allows for non-linear inputs and a set of reachable sets to be computed. The reachable set is the zero sub-level set of the viscosity solution of a time-dependent Hamilton-the solution to the  partial differential equation. There are different ways of formulating this problem and exacting accurate solutions but the afore-mentioned method allows for better generalizations to applications that may be discretized. One drawback of reachability analysis is its computational complexity and corresponding inability to scale well to higher dimensional systems.  Also, the algorithm would end up being extremely cautious. The autonomous vehicle would have a tendency to stay in the same state repeatedly to avoid collisions at all costs. It would end up sacrificing its ability to actually reach its target state. 

Reinforcement learning algorithms  have proven effective in these cases involving navigation by providing suitable trade-offs between optimality and completeness. They are goal oriented wherein the agent maximizes an implicitly or explicitly defined reward function  over a number of steps. Jamie Fisac et. al proposed an effective means of combining Hamilton-Jacobi Reachability and reinforcement learning in his paper in 2019. He demonstrated its success through a variety of simulated robotics tasks and its ability to scale to systems of up to 18 dimensions. It opened up an entirely different avenue in terms of safety analysis in the reinforcement learning domain.   However, his approach was limited to a system with a single agents.

We drew from this baseline and set up a dual agent system involving cyclist interactions for the purpose of this study. 
The database deployed is the Safety Pilot Michigan database which has a large percentage of cyclist events. We formulate the cyclist-vehicle problem as a dual agent zero sum differential game. The autonomous vehicle is the control input, while the disturbance signal is the cyclist.  A noteworthy addition to existing work is that we introduce a means of modelling the disturbance input as function of the cyclist's latent response to the autonomous vehicle's state. The value function is computed using reachability analysis for a three-dimensional system. It is weighted and fed to the deep Q network as a reward that represents the inherent safety factor of a particular state. The deep Q network undergoes training until convergence.

\section{Proposed approach}
Through the proposed navigation algorithm, we aim to provide a safe and optimal means for vehicles to interact with cyclists. 
The following section delineates the dynamics of the system, safety quantification through reachability, effectively modelling disturbance input and finally deep Q learning for navigation.

\subsection{Baseline: Reachability for safety analysis}
This problem involves the formulation of three different sets of states each corresponding to separate levels of safety. The first step towards solving this problem is the definition of the collision set. The states in this set represent collisions in real life between the cyclist and the vehicle. The collision set $C_{0}$ is closed and is represented by
 $g:R^{n} \rightarrow R$
 
 \begin{equation}
 C_{0} = \{s\in R^{n} | v(s) <=0\}
\end{equation}
where $g(s)$ can also be thought of as the minimum reward achieved over time by a trajectory starting in state $x$ accounting only for the best possible control input at every time instant.

The next step involves computing a backward reachable set building from the collision zone. This set is the zero level sub-set of the terminal value time-dependent Hamilton-the solution to the  partial differential equation.  It is representative of a set of  potentially unsafe states that could eventually result in collision. The backward reachable set is therefore
 \begin{equation}
 \begin{split}
 B =\{s\in R^{n}|\exists T\in [-T,0]: \forall u(\cdot) \in U, \forall d(\cdot) \in D:\\ \epsilon(t:s, u(\cdot), d(\cdot))\in C\}
\end{split}
 \end{equation}
 where $\epsilon(t:s,u(\cdot),d(\cdot))$ denotes a trajectory in the system at time t.  The flow field $f:R_{n} \times A \times B \rightarrow R^{n}$is uniformly continuous, bounded, and Lipschitz continuous in $s$.
 
In order to effectively quantify the degree of safety of each state in the backward reachable set, we compute the value function associated with each state for the afore mentioned optimal control problem. The value function $v :R^n \times [-T,0] \rightarrow R$ is representative of the maximum penalty a trajectory accumulates over time despite the best possible control input. It is a cumulative measure of how close the trajectory comes to entering the collision set over time. 
The solution to this problem is given by:
 \begin{equation}\label {val}
 \begin{split}
     min\Bigl\{g(s)-v(s,t), D_{t}v(s,t) + H(s, D_{s}v(s,t))\Bigr\} = 0\\
     v(s,0) = g(s)
 \end{split}
 \end{equation}
 where the resulting Hamiltonian is:
 \begin{equation}
     H(s, p) = \max_{u\in U}\min_{d \in D} p^{T} f(s, u, d) 
 \end{equation}
 
 The complement of the union of the backward reachable set and the collision set is the set $R$  of states that is extremely safe with no chance of collisions in the given time horizon. This set is the robust solution to the optimal control problem in question. If the vehicle enters any state in the set, it will never end up in a collision. However, this also tends to make the navigation algorithm overtly cautious. It is therefore combined with reinforcement learning to enhance optimality of the solution in terms of time taken.
 \begin{equation}
     R = (B \cup C)'
 \end{equation}
 
\subsection{Our approach}
Fisac et. al 2019 proposed the afore mentioned method for safety analysis for a system consisting of a single agent. We extended this work to account for systems with multiple agents. The problem is formulated as a dual agent zero sum differential game where the autonomous vehicle is the ego agent and the cyclist is the disturbance input. The system is modelled as an ordinary differential equation as follows:
\begin{equation}
    \frac{ds}{dt} = \dot{s} = f(x,u,d)
\end{equation}
\begin{equation}
    \dot{s} = \frac{d}{dt}\begin{bmatrix}
\Delta x\\
\Delta v\\
\Delta y
\end{bmatrix} = \begin{bmatrix}0&1&0\\0 &0&0\\0&0&1 \end{bmatrix}\begin{bmatrix}
\Delta x\\
\Delta v\\
\Delta y
\end{bmatrix} + \begin{bmatrix}
0\\
d-u\\
0
\end{bmatrix} 
\end{equation}
 Assume $s\in R^{n}$. Here $s$ is the state of the system, $u(\cdot)$ is the control input of the autonomous ego vehicle and $d(\cdot)$ is the disturbance input of the cyclist. The state $x$ is basically an array containing the longitudinal range $\Delta x$,  longitudinal range rate $\Delta v$ and lateral range $\Delta y$ of the cyclist with  respect to autonomous vehicle.

Assume $u(\cdot) \in U(t)$ and $d(\cdot) \in D(t)$ where
\begin{gather*}
    U(t) = \{\phi:[t,0] \rightarrow A|\phi(\cdot) \text{is measurable}\}\\
    D(t) = \{\phi:[t,0] \rightarrow B|\phi(\cdot) \text{is measurable}\}
\end{gather*}
 where $A\subset R^{n_{u}}$ and $B\subset R^{n_{d}}$ are compact  and $t\in [-T, 0]$ for some $T>0$.
 The solution to the problem involves solving a terminal value time-dependent Hamilton-Jacobi-Bellman partial differential equation.

 \subsection{Main contribution: Modelling disturbance input as a reflection of the cyclist's comfort level}
Existing literature failures  to account for the fact that cyclists and pedestrians are bound to demonstrate feelings of comfort or discomfort with regards to agents around them.
This novel approach highlights the fact that the disturbance input  of the cyclist may not be effectively represented by just its acceleration input at every instant. We model the disturbance input of the cyclist as a function of features such as acceleration input and the autonomous vehicle's state.  Typically, this disturbance set is a proper sub-set of the original disturbance set. 

To effectively represent the cyclist's perception, we deploy an auto-encoder. Labelled data  consisting of safe and dangerous states is fed into the auto-encoder. The segregation of data into dangerous and safe states is done  on the basis of the Hamilton-Jacobi value function computed in the previous subsection. If the corresponding value function $v(x)$ is positive, there exists a solution by the controller to avoid collisions and the state is classified as safe. If the value function is negative or zero, the collision is inevitable in the future as shown in the equation below:
    \begin{equation}
        v(s) > 0   \implies \text{safe state}
    \end{equation}
    \begin{equation}
        v(s) <= 0   \implies \text{dangerous state}
    \end{equation}
 
The next step involved training a classifier to represent as suitable mapping. Rare event classification poses some serious challenges especially with regards to unbalanced training data. Dangerous states account for only 5 to 10 percent of the existing data. Very often classifiers trained through deep learning are unable to effectively reflect these states and are thus less than suitable for classifying rare events.

The auto-encoder approach for classification is similar to anomaly detection. In anomaly detection, we learn the pattern of a normal process. Anything that does not follow this pattern is classified as an anomaly. For a binary classification of rare events, we used auto-encoders.The process has three separate stages: encoding, decoding and characterization of the target class. We trained the auto-encoder on the data that is labelled safe. Safe states are representative of the normal or desired state of the system. The encoder learns the lower dimensional relationship that exists among these features.

The next part involved reconstruction of features. The entire  data-set inclusive of both safe and dangerous states is fed to the decoder. The mean square difference between the input and the output is used for classification. If the reconstruction error is high, it means that the corresponding state was dangerous and vice-versa for safe states. The corresponding mapping $w$ is representative of the cyclist's response to the autonomous vehicle and is fed into the system for the computation of the afore mentioned backward reachable set.

\begin{equation}
    w:(\Delta x, \Delta v, \Delta y, \Delta a) \implies [target, outlier]
\end{equation}
 
\subsection{Deep Q Learning for navigation}
\label{marker}
The solution to the optimal control problem above results in a set of states that are deemed extremely safe. This solution is robust, but it also has a tendency to be overtly cautious. This is undesirable in cyclists vehicle interactions because the optimality of the vehicle's trajectory in terms of time is sacrificed as a result of it being extremely safe. A suitable trade-off between time and safety needs to be reached.

A deep Q network from existing literature with experience replay is deployed. The problem is formulated as a discrete-time Markov process that aims at maximizing the cumulative payoff of a trajectory exponential discounted over time. 
The value function deployed is an extension of the value function computed in [\ref{val}] in its discretized form. 
The policy or the next action is determined by the maximum output of the Q-network
The loss function here is mean squared error of the predicted Q-value and the target Q-value. Learning is carried out using the gradient descent algorithm as shown in the equation \ref{update} for $k = 1, 2, ...$ till convergence.
\begin{equation}\label{update}
  \theta_{k+1} \leftarrow \theta_{k} - \alpha\Delta_{\theta}E_{s'\sim P(s'|s,u)}[(Q_{k}(s,u) - target(s'))^2]|_{\theta = \theta_{k}}
\end{equation}
\begin{equation}\label{eq: loss}
  target = R(s, u, s') + \gamma \max_{u'} Q_{k}(s', u')
\end{equation}

\section{Experimental set-up}

\subsection{Extraction of cyclist events}

The database used is a naturalistic database called the Safety Pilot Michigan Database \cite{feng2018drivers}. It has 34 million miles of driving data logged over four years in the Ann Arbor area. The deployment included approximately 2,800 equipped vehicles and 30 roadside equipment. The data was logged using a Mobileye sensor with a range of 98m and a WSU (Wireless Safety Unit).  This approach deploys annotated data from three  tables: \emph {DataFrontTargets}, \emph {DataWSU} and \emph {DataLane}. 

Typically, we define a cyclist event as a situation spanning the time the Mobileye sensor first detects a particular cyclist to the time the same cyclist exits the sensor's field of vision. The primary data used in the detection of a cyclist event is longitudinal range, longitudinal range rate and lateral range of the ego vehicle with respect to the cyclist. Around 40539 cyclist events were detected. The next step  data involved filtering out false positive events which corresponded to any of the categories listed below:
\begin{enumerate}
\item {An event in which the maximum longitudinal range logged was over 50m. This implied no actual interaction between the cyclist and the vehicle.}
\item {An event that lasted less than 1 second. This implied  noisy sensor data or an interaction of no significance}
\item {An event in which the cyclist is detected on right side of the road: This is beyond the scope of this study}
\end{enumerate}

\subsection{Mapping  disturbance input}
The data-set deployed to model the cyclist's comfort level is the Safety Pilot Michigan Database. Interactions of drivers with cyclists can be analysed as  sequences of states and corresponding actions which progress over time. The state $x$ of the system in this case is a tuple consisting of longitudinal range, longitudinal range rate , lateral range and acceleration  input $\Delta a$ of the cyclist.  All the values are normalized. Studying the entire interaction over the length of its duration and effectively quantifying disturbance input is somewhat implausible. Each state is classified as either safe or dangerous based on the value function as mentioned in the previous section. The SPMD data was not sufficient to construct a suitable classifier. Synthetic states  are fabricated in addition by modifying time to collision and acceleration input of the cyclist. The safe states are fed into the network during the encoding phase. During reconstruction, a mixture of safe and dangerous states is fed into the network. This results in a mapping on the basis of the magnitude of the reconstruction error.

\subsection{Computing the backward reachable set}

Computing a backward reachable set for every possible state and corresponding control input on behalf of the vehicle is extremely time-consuming. To reduce computational complexity,the state of the vehicle is described relative to the cyclist. The state is basically an array containing the longitudinal range $\Delta x$, longitudinal range rate $\Delta v$ and lateral range $\Delta y$ of the cyclist with respect to autonomous vehicle. It is pertinent to note that the the state of the system is computed from the boundary of the collision zone. The collision zone in this case is a circle of 1m  radius. This allows for a one time computation of the reachable set. Ian Mitchell’s level set toolbox is used as the baseline for computation of the backward reachable set. Input to the modified algorithm was the state of the system and the disturbance mapping.

 \subsection{Setting up the deep Q network}
The state of the system is representative of the longitudinal range, longitudinal range rate and lateral range of the cyclist with respect to the vehicle. It also reflects lane boundaries and the distance of the vehicle from the goal position. The action space of the autonomous vehicle is  discretized to represent several combinations of velocities and yaw angles. 

This deep Q network also allows for experience replay. It allocates memory to store experiences and observations of the past. At specific instances during training of the model, a batch of experiences is re-sampled and fed into the neural network. This ensures greater efficacy of the agent in the long run. Correlation is broken and the action values do not oscillate or diverge catastrophically. Adam optimizer is utilized.

\section{Results}

To verify the proposed approach, we conducted two sets of experiments. The first set involved comparing the level set computed through our framework with the one deployed by Fisac et al.  The second set of experiments involved evaluating the test trajectories qualitatively and quantitatively.
 
\begin{figure}
\centering
\begin{subfigure}[a]{0.5\textwidth}

      \includegraphics[width = 1\linewidth, height = 130pt]{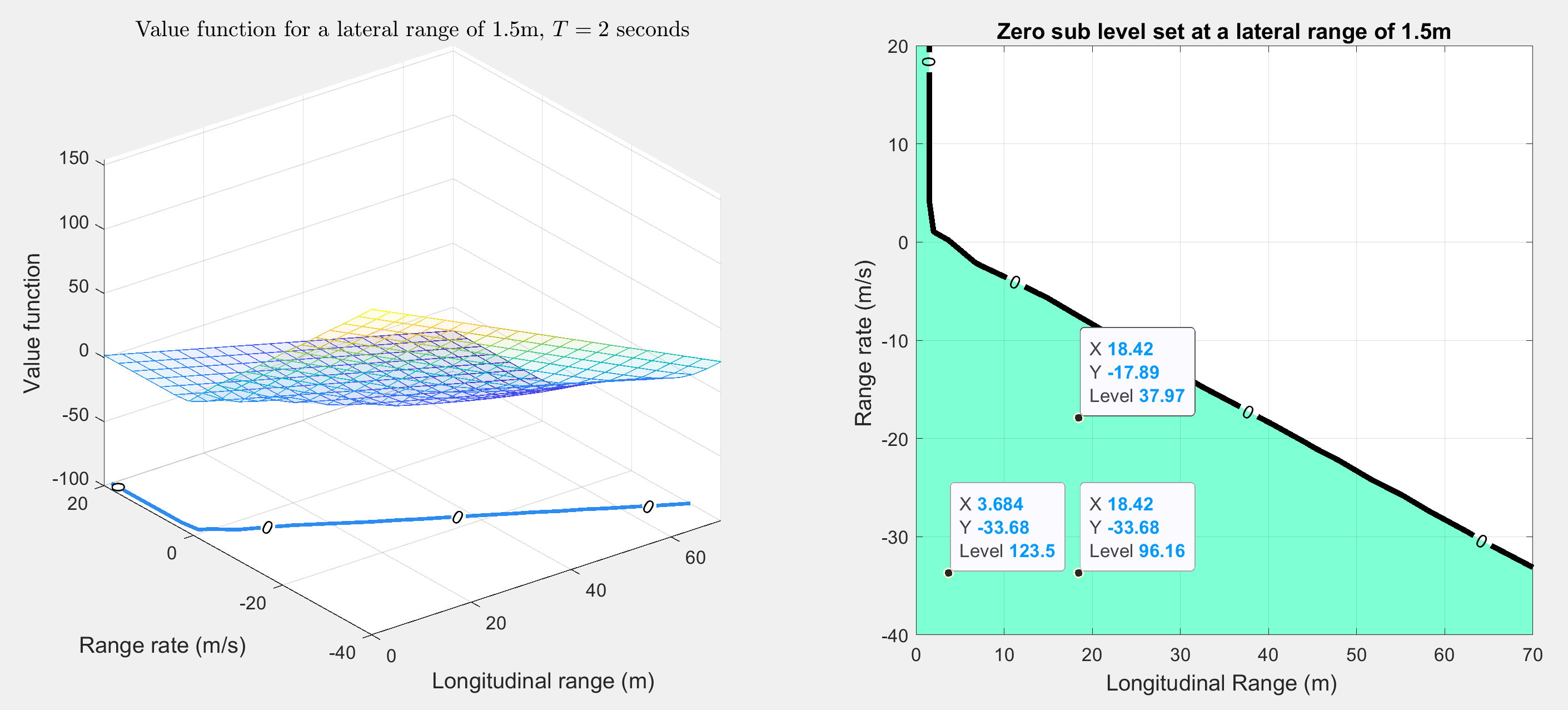}
      \caption{}

\end{subfigure}
\begin{subfigure}[b]{0.5\textwidth}
     
      \includegraphics[width = 1\linewidth, height = 130pt]{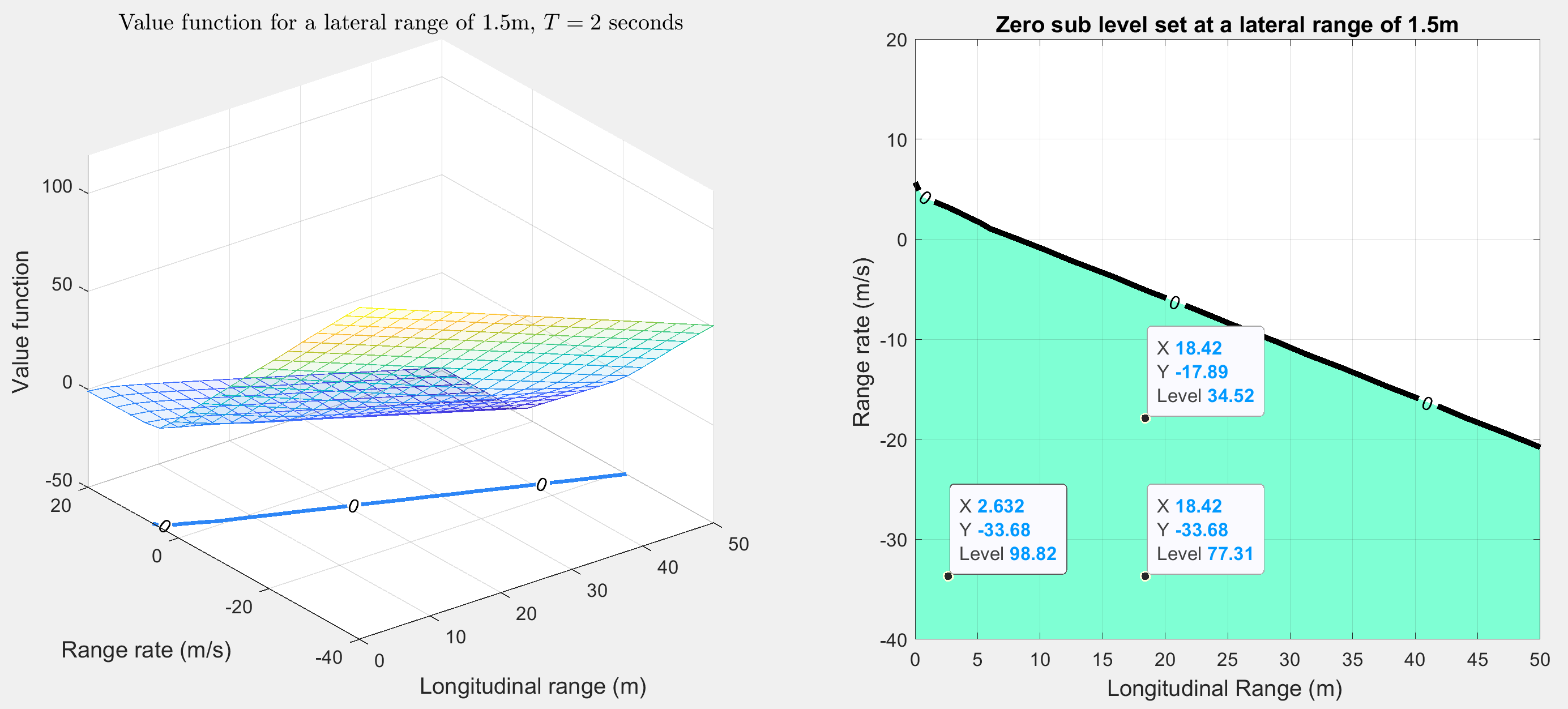}
      \caption{}

\end{subfigure}

\caption{The value function has been plotted on the left side.For the sake of visualization, lateral range has been restricted to a value of 1.5m. On the right side is the zero sub-level set of the viscosity solution to the Hamilton-Jacobi Bellman partial differential equation. In each of the annotated boxes, $X$ represents the longitudinal range, $Y$ represents the longitudinal range rate and $Level$ denotes the negated value function. The aqua green portion represents the backward reachable set (a) Value function computed using our algorithm (b) Value function computed using the algorithm deployed by Fisac et al. 2019}
\end{figure}
\begin{table*}[h]
\caption{Quantitative evaluation of trajectories on the Safety Pilot Michigan Database.  }

\label{table1}
\begin{center}
\begin{tabular}{|p{2.5cm}|p{2.3cm}|p{2.5cm}|p{2.3cm}|p{2.3cm}|p{2cm}|}
\hline
 &Safety Factor &  Unsafe States (\%) & Time (sec) & Collisions & Goal Reached\\
\hline
Human driver & -2.31 & 39.0 & 253 & 0 & 1\\

\hline
Fisac '19 &-2.18 & 36.8 & \textbf{199} & 0 & 1\\
\hline
Our Framework &\textbf{-2.01} & \textbf{31.6} & 233 & 0 & 1\\
\hline

\multicolumn{6}{l}{Bold text is indicative of the best value corresponding to a particular performance metric. The test data-set consists of 60 trajectories.}
\end{tabular}
\end{center}

\end{table*}

\begin{figure*}
      \centering
      \begin{center}
      \includegraphics[width = 1\linewidth]{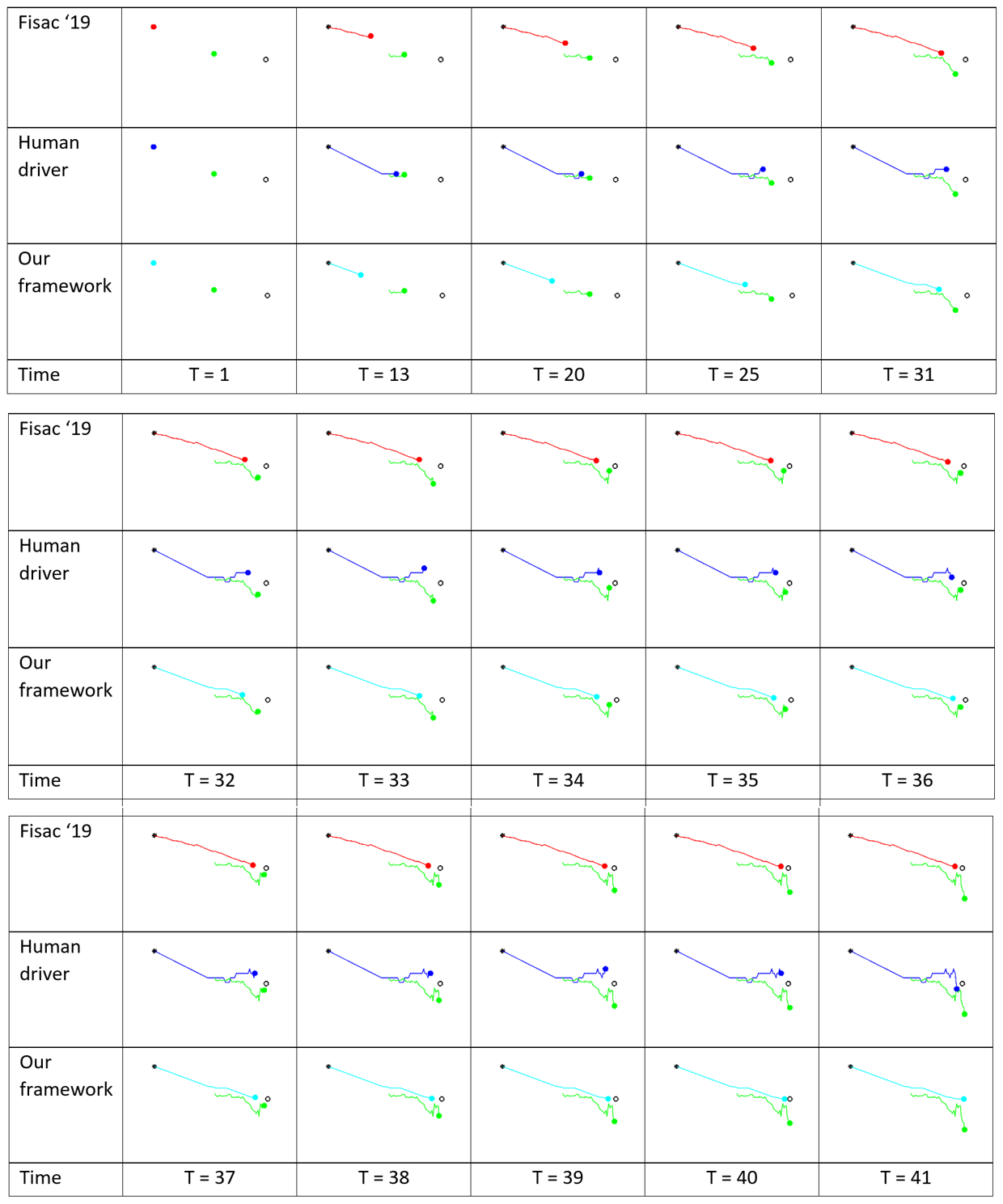}
      \caption{A visualization of a cyclist event from the test set. A red line denotes the trajectory computed using the level set deployed by Fisac et al. A dark blue line denotes the ground truth i.e. the behavior of human drivers. A cyan blue line indicates the trajectory traversed by our framework. A green line denotes the trajectory of the cyclist. The position of each of these agents at a particular time instant is given by a dot on its trajectory in its corresponding color. The start and goal positions are denoted by a black star and black circle respectively }
      \label{fig1}
      \end{center}
\end{figure*}

\subsection{Evaluating the level sets and corresponding value function}
In their paper published in 2019, Fisac et al. demonstrated success when it came to training a single agent using deep Q networks and reachability. We extended this framework to account for a system with two agents and computed the corresponding value functions. We used this level set as a baseline for comparison to the level set we computed by modifying the disturbance input.

The optimal control problem in question has three dimensions. But for the sake of visualization, a slice of the value function is represented in figure 1 at a lateral range of 1.25m. The zero level subset is shown in. Upon observation of the the annotated values in the two images, it is evident that the value functions computed are an effective reflections of the safety of the state. 

After carefully comparing the two images, we notice an extra portion of states in the potentially unsafe region corresponding to a longitudinal range of around 2m and a range rate of over 2m/s. This extra portion is the result of the modifications we made to the disturbance input to reflect the reaction of cyclists. These states basically correspond to a scenario where the autonomous vehicle is quite close to the cyclist; but is operating at a much lower velocity than the cyclist. In theory, this would not be an unsafe interaction. But in practice, the cyclist would perceive these states as a threat merely because the autonomous vehicle is too close to it. 

We can therefore say that our approach has been able to successfully reflect the perception of humans which is one of the main goals of autonomous navigation. 

\subsection{Evaluation of trajectories}
We evaluated our algorithm on 60  cyclist events from the Safety Pilot Michigan Database. It is pertinent to note that these events have not been sampled during training. We use five  quantitative metrics for comparison as shown in table \ref{table1}:
\begin{enumerate}

\item{\textit{Unsafe states encountered}: This is the ratio of the number of times a state in the backward reachable set is encountered to the total number of states encountered. It indicates the number of times the vehicle finds itself in a potentially unsafe region  }
\item{\textit{Safety factor}: This is the  average Hamilton-Jacobi value function per trajectory to the total number of trajectories in the test set. A lower value indicates a trajectory is more unsafe}
\item{\textit{Optimality}: This is the cumulative sum of the time taken to complete all the trajectories in the test set given each of their start positions and goal positions }
\item{\textit{Collision}: This is the number of times the vehicle finds itself within a 1 m radius of the cyclist during testing}

\item{\textit{Goal Reached}: This is the ratio of the number of times the vehicle reached the goal to the total number of events in the test set.}
\end{enumerate}
We compare our algorithm with the one deployed by J Fisac et al. in 2019 and also against the trajectories followed by human drivers from the SPMD Database. Our algorithm supersedes both human driver trajectories and Fisac'19's agent when it comes to safety. While it takes less time on an average to navigate around cyclists as compared to the ground truth, it takes slightly more time than the algorithm deployed by Fisac '19. This can be attributed to the fact that the level set computed in Fisac '19 fails to reflect the level of comfort the cyclist feels during the interaction.

A visualization of one interaction from the test set is shown in figure \ref{fig1}. In this particular interaction, our algorithm works best in terms of time taken, completeness and safety analysis. It reaches the goal state faster than the human driver and Fisac 19's agent. Its path is proven optimal in terms of length from the visualization. Its safety factor is also particularly high.

\section{CONCLUSIONS}
This paper successfully demonstrates the use of deep Q networks in combination  with the Hamilton-Jacobi Bellman value function to safely interact with cyclists. It is also able to effectively reflect a cyclist's response to the state of the autonomous vehicle in terms of its disturbance input. This can also be interpreted as a really good indicator of the cyclist's comfort. Our framework  performs well in terms of time taken and ability to reach the target when tested on events from the Safety Pilot Michigan Database. Safety analysis is also effectively quantified in this work.
We are currently working on extending this approach to multi-agent systems for navigation.

\section{Mathematical Proofs}

 Let $\mathcal{X}\subset X$ and $\mathcal{U}\subset U$ be finite discretizations of the state and action spaces, let $\mathcal{D}\subset D$, and let $f : \mathcal{X} \times \mathcal{U} \times \mathcal{D} \rightarrow \mathcal{X}$ be a discrete transition function approximating the system dynamics. The proof for deep Q learning follows from the set-up in Section IV-D.

Here, learning rate $\alpha$ is 0.0001. It satisfies the following assumptions
\begin{equation}\label{alpha}
  \sum_{k} \alpha_{k}(s, u, d) = \infty, \quad   \sum_{k} \alpha_{k}^2(s, u, d) < \infty 
\end{equation}
 
\subsection{Contraction mapping}
Proof: For all states $s\in X$, prove that $\mid F[Q] (s) - F[Q'] (s)\mid < k \parallel Q - Q'\parallel_{\infty}$\\

\noindent $\mid F[Q] (s) - F[Q'] (s)\mid$ \\
$= \parallel(R(s, u, s') + \gamma \max_{u'} Q_{k}(s', u')) - (R(s, u, s') + \gamma \max_{u'} Q'_{k}(s', u')) \parallel$\\
$\leq \gamma (\max_{u'} Q_{k}(s', u') - \max_{u'} Q'_{k}(s', u'))$\\

Assume the first maximum is the larger one without loss of generality, assume $u$ achieves this\\
\noindent $\mid F[Q] (s) - F[Q'] (s)\mid$ \\
$\leq \gamma \max_{u'}( Q_{k}(s', u') -  Q'_{k}(s', u'))$\\
$\leq \gamma \max_{u'}( Q_{k}(s', u') -  Q'_{k}(s', u'))$\\
$\leq k \parallel Q - Q'\parallel_{\infty}$

\subsection{Convergence of Q learning under our assumptions}
Our assumptions are aligned with assumptions 1, 2, 3 and 5 in \cite{tsitsiklis1994asynchronous}. This basically means convergence is guaranteed in the case of deep Q learning following from the general proof of Q learning convergence for ﬁnite-state,ﬁnite-action Markov decision processes presented in [29].

\addtolength{\textheight}{-12cm}   





.

{
\bibliographystyle{IEEEtran}
\bibliography{egbib}
}

\end{document}